%% file: root.tex
\documentclass[letterpaper, 10 pt, conference]{ieeeconf}  

\IEEEoverridecommandlockouts                              

\usepackage{cite}
\usepackage{amsmath,amssymb,amsfonts}
\usepackage{bm}
\usepackage[caption=false, font=footnotesize]{subfig}
\usepackage{algorithmic}
\usepackage{graphicx}
\usepackage{textcomp}
\usepackage{xcolor}
\usepackage{tikz}
\usepackage[symbol,hang,flushmargin]{footmisc}
\usepackage[normalem]{ulem}
\usepackage[nolist]{acronym}
\acrodef{spi}[SPI]{Shadow Program Inversion}
\acrodef{dpse}[$\partial$PSE]{Differentiable Programming in Stochastic Environments}
\acrodef{tcp}[TCP]{tool center point}
\acrodef{pca}[PCA]{Principal Component Analysis}
\acrodef{nsga}[NSGA-II]{nondominated sorting genetic algorithm}
\acrodef{tht}[THT]{through-hole technology}
\acrodef{pcb}[PCB]{printed circuit board}
\acrodef{gmm}[GMM]{Gaussian Mixture Model}
\acrodef{maml}[MAML]{model-agnostic meta-learning}
\acrodef{fomaml}[FOMAML]{first-order MAML}
\acrodef{dp}[$\partial$P]{Differentiable programming}
\acrodef{nn}[NN]{neural network}
\acrodef{tl}[TL]{Transfer learning}

\newcommand\copyrighttext{%
  \footnotesize \textcopyright 2022 IEEE. Personal use of this material is permitted.
  Permission from IEEE must be obtained for all other uses, in any current or future 
  media, including reprinting/republishing this material for advertising or promotional 
  purposes, creating new collective works, for resale or redistribution to servers or 
  lists, or reuse of any copyrighted component of this work in other works.}
\newcommand\copyrightnotice{%
\begin{tikzpicture}[remember picture,overlay]
\node[anchor=south,yshift=10pt] at (current page.south) {\fbox{\parbox{\dimexpr\textwidth-\fboxsep-\fboxrule\relax}{\copyrighttext}}};
\end{tikzpicture}%
}

\title{\LARGE \bf
Heuristic-free Optimization of Force-Controlled Robot Search Strategies in Stochastic Environments
}

\author{Benjamin Alt$^{1,2}$, Darko Katic$^{1}$, Rainer Jäkel$^{1}$ and Michael Beetz$^{2}$%
\thanks{This  work  was  supported  by  the  German  Federal  Ministry  of  Education and Research under the grant 01DR19001B.}
\thanks{\raggedright$^{1}$ArtiMinds Robotics, Karlsruhe, Germany {\tt\footnotesize benjamin.alt@artiminds.com}}%
\thanks{$^{2}$Institute for Artificial Intelligence, University of Bremen, Germany}%
}

\begin{document}

\maketitle
\copyrightnotice
\thispagestyle{empty}
\pagestyle{empty}

\begin{abstract}
In both industrial and service domains, a central benefit of the use of robots is their ability to quickly and reliably execute repetitive tasks. However, even relatively simple peg-in-hole tasks are typically subject to stochastic variations, requiring search motions to find relevant features such as holes. While search improves robustness, it comes at the cost of increased runtime: More exhaustive search will maximize the probability of successfully executing a given task, but will significantly delay any downstream tasks. This trade-off is typically resolved by human experts according to simple heuristics, which are rarely optimal. This paper introduces an automatic, data-driven and heuristic-free approach to optimize robot search strategies. By training a neural model of the search strategy on a large set of simulated stochastic environments, conditioning it on few real-world examples and inverting the model, we can infer search strategies which adapt to the time-variant characteristics of the underlying probability distributions, while requiring very few real-world measurements. We evaluate our approach on two different industrial robots in the context of spiral and probe search for THT electronics assembly.\footnote[1]{See \texttt{github.com/benjaminalt/dpse} for code and data.}
\end{abstract}

\section{Introduction}
\label{sec:introduction}
Despite years of research, manipulating objects whose pose can only be determined with uncertainty is still very challenging for robots. In service robotics, examples include the manipulation of occluded objects or imprecise pose estimation due to noisy perception. Industrial applications often require robots to manipulate objects whose pose in the workspace is known only within given tolerances and whose precise pose varies stochastically between cycles. In the context of electronics assembly, for example, the poor positioning accuracy of conveyor belts, manufacturing tolerances of parts from different suppliers or wear and tear of materials make it impossible to precisely know the pose of connectors or holes during offline programming \cite{metzner_high-precision_2021}. Force-controlled search strategies, which probe the environment with a defined series of motions until a change in forces is detected, are a commonly used method for resolving such uncertainties \cite{jasim_position_2014}. While greatly increasing the robustness of robot tasks, however, force-controlled search comes at the cost of increased time required to complete the task. Determining an optimal set of search motion parameters, which maximizes robustness while keeping execution times minimal, currently requires lengthy parameter tuning by human robot programmers. We posit that a method for the automatic parameterization of search strategies must meet four crucial requirements to be applicable for most real-world use cases:
\begin{enumerate}
	\item \textit{Heuristic-freeness:} It must make no prior assumptions about the underlying stochastic processes.
	\item \textit{Data efficiency:} It must require only unsupervised or self-supervised data. The amount of real-world training data must be small and data collection must not interfere with the completion of regular tasks, such as the ongoing operation of a production line.
	\item \textit{Handling nonstationary processes:} It must be robust to nonstationary noise processes whose characteristics change over time.
	\item \textit{Black-box optimization:} It must be agnostic with respect to the concrete robot program representation to ensure cross-domain applicability and compatibility with various robot frameworks.
\end{enumerate}

\begin{figure}
	\includegraphics[width=\linewidth]{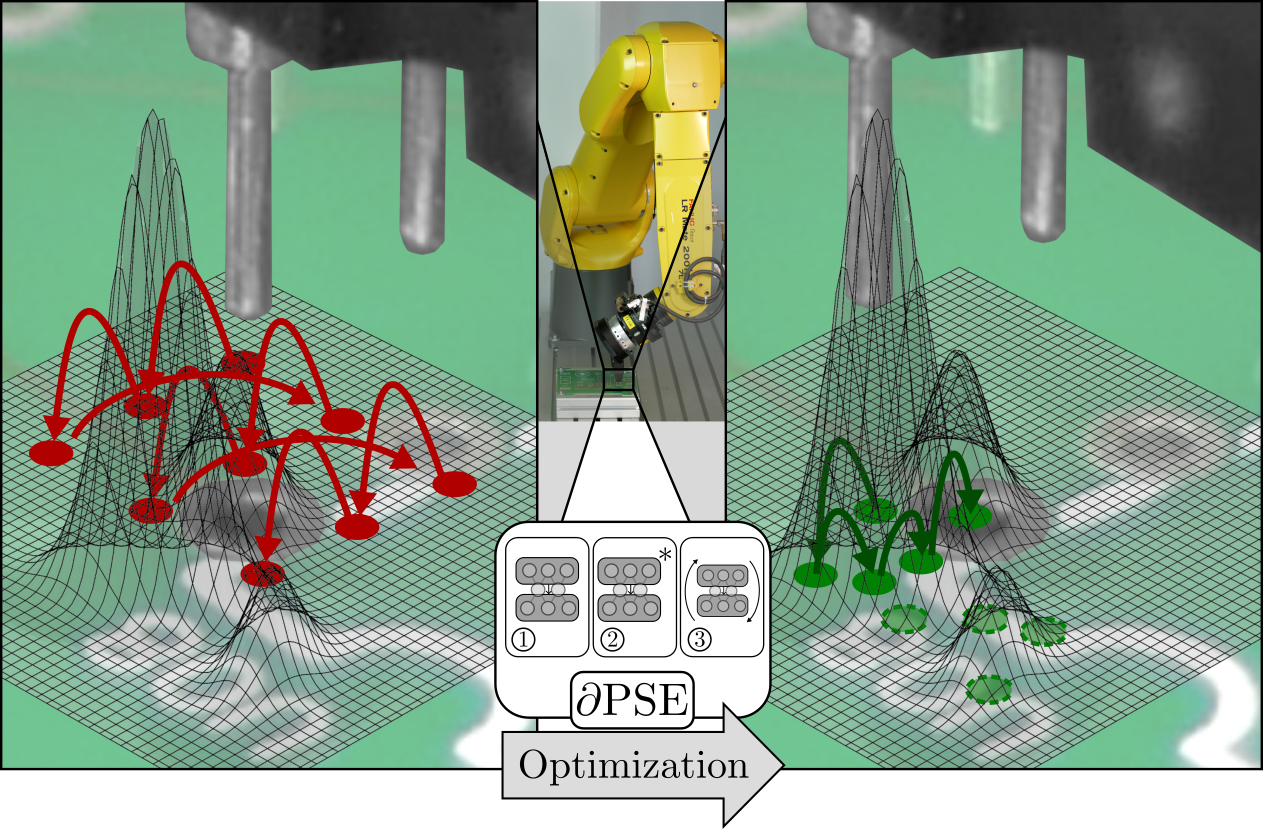}
	\caption{\acf{dpse} leverages transfer learning and differentiable robot program representations to optimize search motions in the presence of stochastic noise.}
	\label{fig:overview}
\end{figure}

In this paper, we introduce \acf{dpse}, a heuristic-free, data-efficient black-box approach to optimizing robot search motions even in the face of nonstationary stochastic processes.

Our contribution is threefold:
(1) A method for \textit{learning predictive models of robot skills} which capture the stochastic characteristics of the environments they are executed in. To that end, we present a transfer-learning based approach to efficiently condition such models on the stochastic environment at hand. By pre-training in simulated environments and fine-tuning on a small dataset of real-world action executions, the learned models predict an expected robot trajectory given the robot skill's parameters (e.g. the search pattern of a probe search) and the (learned) probability distributions governing the environment. We show that continuous fine-tuning can keep the learned models synchronized with time-varying, nonstationary stochastic environments.\\
(2) A method to \textit{optimize motion parameters} to ensure minimal search time and maximal roboustness in the face of nonstationary stochastic environments. We embed these environment-conditioned models in a state-of-the-art framework for differentiable robot programming \cite{alt_robot_2021}, in which robot programs can be expressed as differentiable computation graphs. A gradient-based optimizer jointly optimizes the program's input parameters with respect to relevant metrics such as cycle time or failure rate.\\
(3) \textit{Experimental validation} of our method in two real-world mechanical and electronics assembly tasks for both stationary and nonstationary environments, and empirical comparison against several meta-learning approaches.

\begin{figure*}[ht!]
	\centering
	\includegraphics[width=\textwidth]{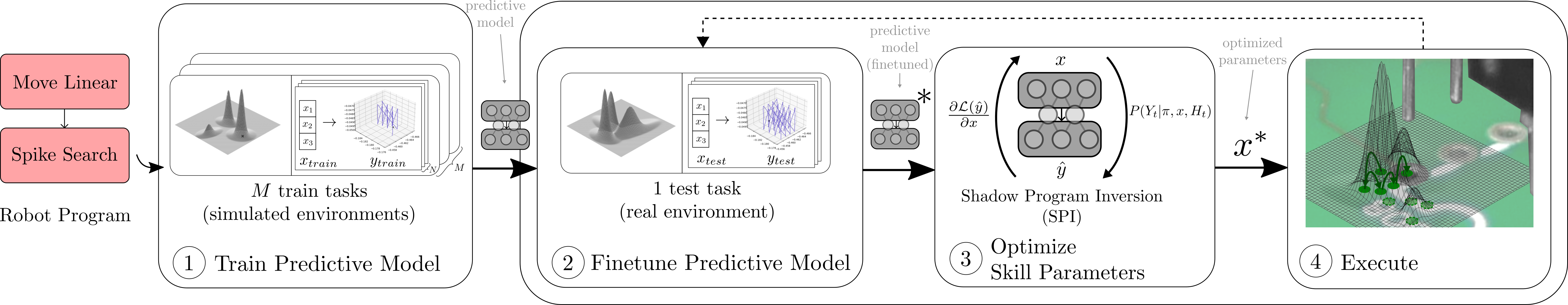}
	\caption{Overview of our proposed approach. \ac{dpse} enables the data-efficient, heuristic-free optimization of arbitrary parametric search strategies by pretraining a predictive model (``shadow program'') of the search strategy on a large simulated dataset (1), finetuning it on a few samples of the concrete real-world environment (2), and inferring optimal search parameters via \ac{spi} (3). By repeatedly finetuning on data collected at test time (4), nonstationary noise processes can be addressed.}
	\label{fig:overview_wide}
\end{figure*}

\section{Related Work}
\paragraph{Robot program parameter optimization} Several black-box optimizers for robot program parameters have been proposed, most often relying on zero-order methods such as evolutionary algorithms \cite{marvel_automated_2009,chernova_evolutionary_2004}, whose requirement
of repeatedly executing the robot program for every optimization
renders them impractical for real-world application. Bayesian optimization \cite{calandra_bayesian_2016, akrour_local_2017, berkenkamp_bayesian_2020} avoids this, but requires the goal function to be known at data collection time, is difficult to extend to nonstationary processes \cite{snoek_input_2014, hebbal_bayesian_2019} and cannot exploit gradient information. First-order \ac{nn}-based optimizers have been proposed \cite{zhou_movement_2020, kulk_evaluation_2011}, but remain impractical in productive environments due to their data requirements and reliance on supervised or reinforcement learning. 

\paragraph{\ac{dp}} In \ac{dp}, programs are composed of differentiable operations, which permits the first-order optimization of a program's parameters with respect to its outputs \cite{baydin_automatic_2018}. PDP \cite{jin_pontryagin_2020} is a \ac{dp} framework for differentiable motion control. However, PDP is not black-box, cannot represent hierarchical programs and the learned parameters become outdated in nonstationary environments. \ac{spi} \cite{alt_robot_2021} is a black-box \ac{dp} framework for representing robot programs, which, like Bayesian optimization, relies on surrogate models learned via unsupervised learning. It exploits gradients for efficient optimization, but assumes stationarity and requires substantial amounts of real-world  data. We leverage \ac{spi} for optimization, but avoid its stationarity assumption and drastically reduce the amount of required real-world data.

\paragraph{\acf{tl}} \acused{tl}\ac{tl} is concerned with improving the performance of systems trained on some source task on different, but related target tasks \cite{zhuang_comprehensive_2020}. In \textit{sequential \ac{tl}} schemes, a \ac{nn} first learns a shared set of features on a large source dataset before fine-tuning on a much smaller target dataset \cite{ruder_neural_2019}. In natural language processing \cite{ziegler_fine-tuning_2020, sun_how_2019, mosbach_stability_2020, liu_roberta_2019, peng_transfer_2019} and computer vision \cite{chen_pre-trained_2021, li_transfer_2020, miglani_skin_2021}, unsupervised pretraining outperforms learning from scratch \cite{erhan_why_2010, zhang_language_2018}, particularly for small target datasets \cite{wang_pretrain_2020}. In robotics, sequential \ac{tl} has been applied for inter-robot learning \cite{dasari_robonet_2019}, Sim2Real adaption \cite{julian_never_2020} or few-shot learning \cite{coninck_learning_2019, yen-chen_learning_2020}. We show that in the context of parameter optimization, sequential \ac{tl} not only greatly reduces the amount of required real-world data, but also that continuous fine-tuning as a form of ``lifelong learning'' enables parameter optimization in nonstationary stochastic environments.

\paragraph{Meta-learning}
Meta-learning is concerned with efficiently learning to solve unseen tasks and has been applied with particular success in few-shot problems \cite{hospedales_meta-learning_2021, goyal_inductive_2021}. While some approaches such as Meta Networks \cite{munkhdalai_meta_2017} or Prototypical Networks \cite{snell_prototypical_2017} propose specific architectures, model-agnostic methods such as MAML \cite{finn_model-agnostic_2017} or Reptile \cite{nichol_first-order_2018} optimize arbitrary networks by learning initial network parameters particularly conducive to test-time fine-tuning. We provide an evaluation of several model-agnostic meta-learners as alternatives to our proposed \ac{tl}-based approach in the context of search strategy parameter optimization.

\section{Parameter Optimization in Stochastic Environments}

\subsection{Definitions and Notation}
As an example for an industrial assembly task subject to stochastic variations, we consider a peg-in-hole task, where the actual pose of the hole can be described by a random variable $H_t$ drawn from the (possibly nonstationary) stochastic process $\{H_t\}$. Because the actual hole pose at time $t$ is uncertain, an industrial robot will require the use of search strategies to perform this task reliably. We define a \textit{search strategy} as a robot program which accepts a vector of parameters $x \in \mathcal{X}$ and executes a sequence of end effector motions until a hole has been found or some other termination criterion was reached. Intuitively, a \textit{robot program} can be defined as a function $\pi$ mapping some program parameters $x$ to the \ac{tcp} trajectory $y$ resulting from executing it. However, because this trajectory depends on the uncertain pose of the hole, we model it instead as a random variable $Y_t$, which takes values $y \in \mathcal{Y}$ according to $P(Y_t | \pi, x, H_t)$, the probability distribution over trajectories resulting from the execution of program $\pi$ at time $t$ with inputs $x$ and hole distribution $H_t$.

\subsection{Parameter Optimization via Differentiable Programming}
In the context of robotics, \ac{dp} approaches center around learning or constructing a differentiable representation of $\pi$, which permits to backpropagate losses over the program's outputs to its inputs $x$. \ac{spi} \cite{alt_robot_2021} approximates $P(Y_t|\pi, x)$ by training an auxiliary, differentiable graph of \acp{nn} (``shadow program'', $\hat{\pi}$) via unsupervised learning from past executions of $\pi$. It then optimizes the program's input parameters by \textit{inverting} the shadow program via gradient descent in the shadow program's input space. This optimization can be performed with respect to some user-defined function $\mathcal{L}$ of the shadow program's output (the expected trajectory $\hat{y}$), such as cycle time. \ac{spi}'s compatibility with most existing robot program or skill frameworks and exclusive reliance on unsupervised learning are highly advantageous in industrial robot applications. \ac{spi} is a \textit{heuristic-free} and \textit{black-box} optimizer according to the criteria outlined in section \ref{sec:introduction}. In its model training phase, however, \ac{spi} requires training tuples $(x_i, y_i)$ which cover the region of the input space $\mathcal{X}$ relevant for optimization. This effectively requires sampling the input space during data collection, which is impractical in a running production line. Moreover, if features of the environment such as the poses of holes follow \textit{nonstationary} stochastic processes, the learned shadow program immediately becomes outdated as the probability distributions change over time.

\subsection{Environment-Conditioned Skill Models}
\label{sec:tspi}
Transfer-learning methods are capable of efficiently adapting learned models to changing data distributions \cite{prapas_continuous_2021}. By treating the model learning phase of \ac{spi} as a \ac{tl} problem and applying a pretraining and finetuning technique, we obtain a heuristic-free black-box differentiable program representation which supports the optimization of search motions in the face of nonstationary noise processes with high data efficiency.\\
We first define a \textit{task} $\mathcal{T} = \big\{\{H_t\}, P(Y_t | \pi, x, H_t)\big\}$: For a stochastic process $\{H_t\}$ generating hole distributions, \ac{spi} must learn the probability distribution over the resulting trajectories when executing program $\pi$ with inputs $x$ in this environment. The corresponding \textit{task dataset} $D_{\mathcal{T}_0} = \{(x_0, y_0), ..., (x_N, y_N)\}$ for a task $\mathcal{T}_0$ consists of input-trajectory pairs collected by executing $\pi$ $N$ times, each time sampling a new hole distribution $H_n$ from $\{H_t\}_0$. Following a pretraining/finetuning regime, we propose to pretrain \ac{spi}'s shadow program on a large \textit{source dataset} $D_S = \bigcup_{m=0}^{M}D_{\mathcal{T}_m}$, the collection of $M$ task datasets, each for a different hole-distribution-generating process $\{H_t\}_m$ (see fig. \ref{fig:overview_wide} (1)). The shadow model can then be finetuned on the much smaller \textit{target dataset} $D_T=\{D_{\mathcal{T}_{curr}}\}$ (see fig. \ref{fig:overview_wide} (2)), containing only data from the current task $\mathcal{T}_{curr}$ - in the case of a running production line, the input-trajectory pairs from the last $N$ executions of the program. The core intuition is to learn a prior over many possible environments (hole-distribution-generating processes, $\{H_t\}_m$) offline and finetune on the concrete process at hand.

\subsubsection{Data efficiency}
We found that for most peg-in-hole tasks under uncertainty, only $N$=128 samples per task are required to effectively finetune on the concrete task. Moreover, we show that pretraining on a large ($M$=1000, $N$=128) source dataset collected in a simulated environment and finetuning on one single real-world target task bridges the sim-to-real-gap sufficiently for \ac{spi} to perform meaningful parameter optimization in the real-world environment. Both spiral and probe search strategies considered in experiments \ref{sec:spiral_experiment} and \ref{sec:spike_experiment} involve force-sensitive interactions with the environment; in both cases, pretraining exclusively on simulated data sufficed to find near-optimal parameterizations in the real world. This efficiency can be attributed to the fact that after pretraining, the shadow program can rely on useful priors at two levels: The differentiable motion planner at the heart of the \ac{spi} shadow program architecture provides a strong prior for the expected trajectory in an ideal environment, and the new pretraining step trains a prior over a wide range of possible environments, providing a very good initialization of the shadow program for finetuning.
\subsubsection{Passive data collection}
As a corollary of these strong priors, the proposed pretraining/finetuning regime avoids the need to sample the parameter space in the real production line. Because the relationship between program inputs and outputs is learned across a wide variety of environments and for inputs $x$ sampled from the entire parameter space, the finetuning dataset $D_T$ does not require such diversity anymore. In fact, we find that \ac{spi} still performs meaningful parameter optimization even if all finetuning examples in $D_T$ contain the same inputs, i.e. $D_T=\{D_{\mathcal{T}_{curr}}\}=\{(x_0, y_0), ..., (x_0, y_N)\}$. Consequently, the required 128 real-world finetuning examples can be collected in a completely passive manner by simply collecting robot trajectories from the running assembly line, without needing to sample a diverse set of program parameters and potentially disrupting production.
\subsubsection{Continuous learning and optimization for nonstationary processes}
Nonstationary processes require continuous re-optimization of program parameters, ideally at every timestep $t$. With the proposed pretraining/finetuning regime, this becomes straightforward: After finetuning once, running \ac{spi}'s optimization step and updating the program with the optimized parameters, the next trajectory at timestep $t+1$ can be observed and added to the finetuning dataset. \ac{spi}'s shadow program can then be finetuned again and the optimal parameters for timestep $t+2$ can be computed. Repeating this cycle of finetuning the shadow program and optimizing program parameters ensures that the shadow program's estimation of $P(Y_t | \pi, x, H_t)$ does not deteriorate as $t$ increases. We implement $D_T=\{D_{\mathcal{T}_{curr}}\}$ as a ring buffer of fixed size $N$=128 to ensure that outdated data is eventually removed from the finetuning dataset. To avoid catastrophic forgetting, we finetune the original pretrained model at each iteration instead of repeatedly finetuning the finetuned model.

\begin{figure*}[ht!]
	\subfloat[a]{%
		\includegraphics[width=.2\textwidth]{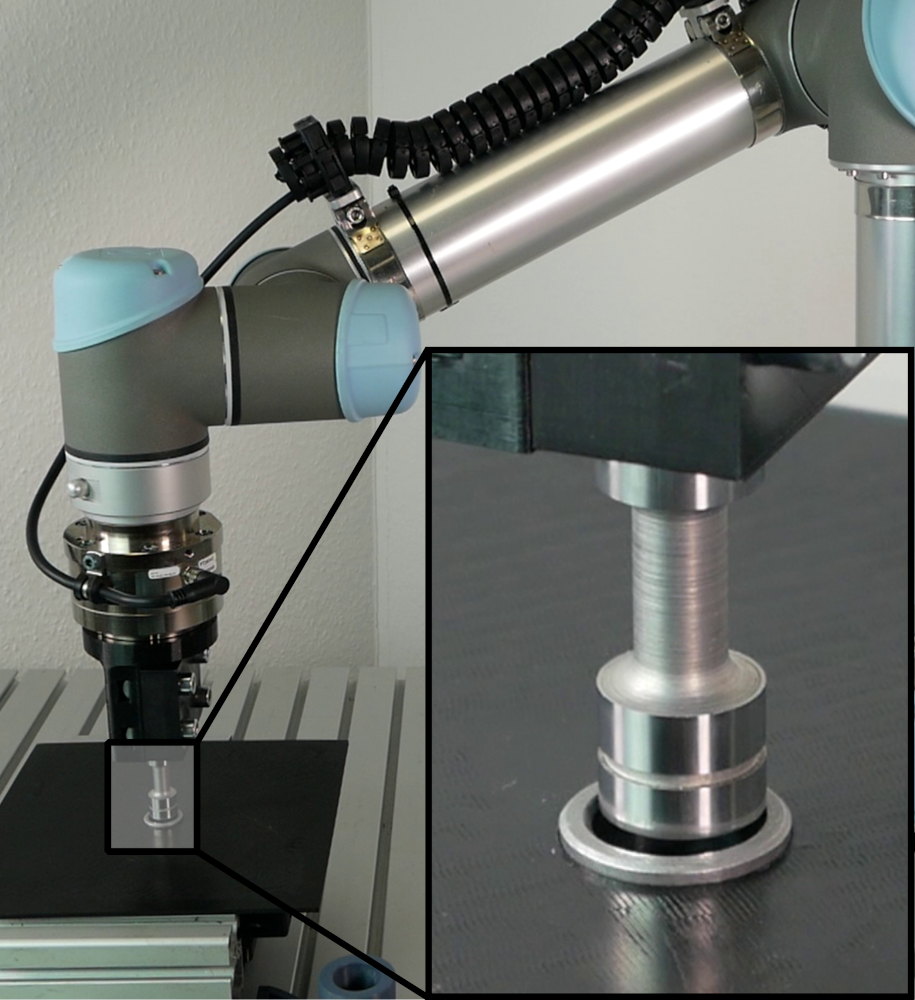}
	}%
	\subfloat[b]{%
		\includegraphics[width=.3\textwidth]{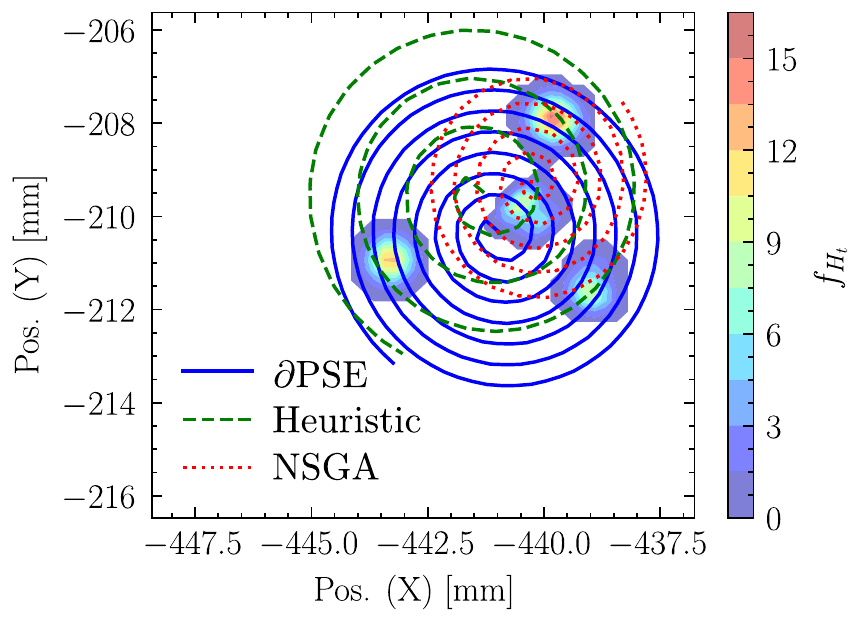}
	}%
	\subfloat[c]{%
		\includegraphics[width=.5\textwidth]{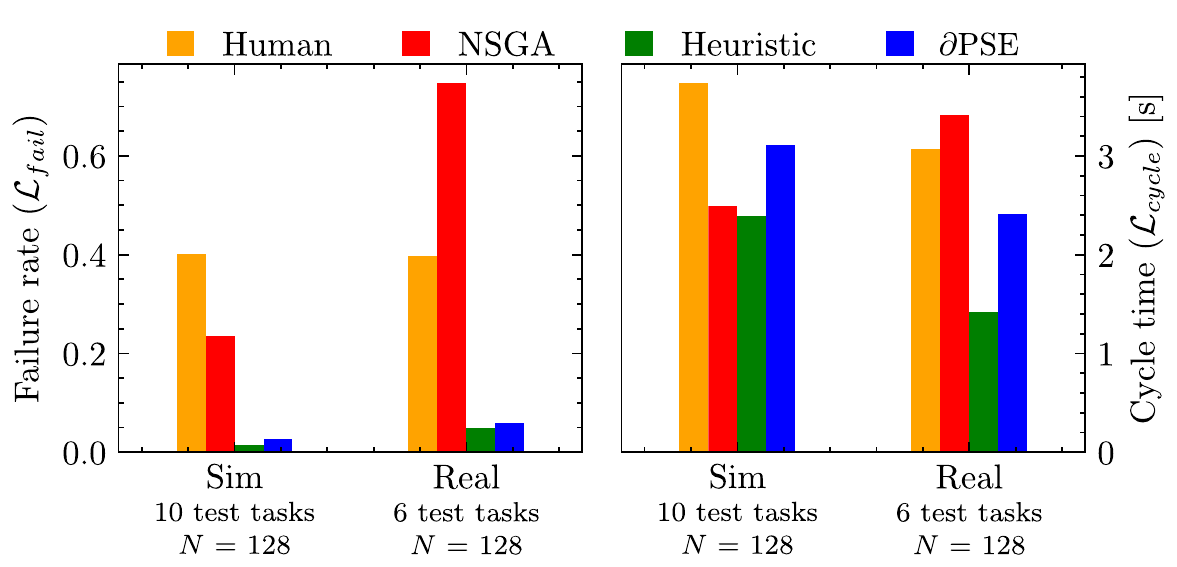}
	}%
	\caption{Results for experiment \ref{sec:spiral_experiment} (spiral search, stationary): Experiment setup (a) and examples for an optimized search pattern generated by \ac{dpse} (b). Heuristic-free \ac{dpse} outperforms the human and \acs{nsga} baselines and is competitive with an application-specific heuristic (c).}
	\label{fig:spiral_experiment_results}
\end{figure*}

\section{Experiments}
We validate our approach by comparing its performance against state-of-the-art baselines on two real-world assembly tasks involving spiral and probe search strategies, respectively. We also evaluate its performance on several simulated nonstationary processes and provide an empirical analysis of our \ac{tl} scheme with respect to possible meta-learning based alternatives.

\subsection{Spiral Search (stationary)}
\label{sec:spiral_experiment}
We consider the mechanical assembly task of inserting a tightly-fitting cylinder into a hydraulic valve (see fig. \ref{fig:spiral_experiment_results}). To simulate realistic process variances, the position of the valve body in the XY-plane is sampled from a bivariate Gaussian mixture with six components, forming the stationary process $\{H_t\} = H_{t} = \sum_{i=1}^{6}w_i\mathcal{N}(\bm{\mu}_i, \bm{\Sigma}_i)$. To successfully perform the insertion, the robot follows a \textit{spiral search} strategy, maintaining a constant pushing force against the surface until dropping into the hole. We seek to optimize the parameters of the search strategy (spiral position, orientation, extents, number of windings, velocity and acceleration) to minimize a linear combination of cycle time $\mathcal{L}_{cycle}(y)$ and failure probability $\mathcal{L}_{fail}(y)$, both of which can be expressed in terms of the expected trajectory $y$ output by \ac{dpse}'s shadow program. \ac{dpse} is pretrained on a source dataset collected by simulating spiral searches on $M_{train}=1000$ different Gaussian mixtures with $N_{train}=128$ sampled hole poses each, and finetuned on $N_{test}=128$ training examples for the $M_{test}=1$ concrete hole distributions at hand.\footnote{Simulations were conducted in a lightweight scripted environment modelling Newtonian physics, damping and friction.} We compare against the following baselines:
\begin{enumerate}
	\item A fixed parameterization by a human expert
	\item A near-optimal \ac{pca}-based heuristic fitting the spiral orientation and extents to the ground-truth valve body poses
	\item $\mu+\lambda$ \ac{nsga} \cite{deb_fast_2002} with $\mu = \lambda = 25$
\end{enumerate}
We repeat the experiment for 10 different distributions $\{H_t\}$ in a purely simulated environment, and for 6 different $\{H_t\}$ on a real UR5 robot.

The results are shown in figure \ref{fig:spiral_experiment_results}. \ac{dpse} provides near-perfect success rates and significantly outperforms \ac{nsga} and the human expert, effectively replicating the results in \cite{alt_robot_2021} with significantly fewer real-world training examples. With respect to cycle time, \ac{dpse} outperforms both the human and \ac{nsga} on the real task. The poor real-world performance of \ac{nsga} illustrates the fundamental disadvantage of most heuristic-free, zero-order black-box optimizers, which they require a large amount of program executions to converge on a good solution. Here, was allowed 250 executions per test distribution, nearly twice \ac{dpse}'s real-world data requirement of 128, suggesting that \ac{dpse} owes its data efficiency at least partly to its exploitation of gradient information.

\subsection{Probe Search (stationary)}
To evaluate \ac{dpse} for a more complex search strategy, we consider the assembly of \ac{tht} electronics components, where the pose of mounting holes on a \ac{pcb} is subject to stationary noise $\{H_t\}$ with an a priori unknown distribution, e.g. from imprecise positioning on a conveyor belt. To avoid scratching the surface of the \ac{pcb}, \textit{probe search} repeatedly touches the surface according to a predefined search pattern until the \ac{tht} component's pins drop into the holes (see fig. \ref{fig:overview_wide} (4)). The search pattern is typically defined as a regular grid, though the search strategy can be significantly optimized by tailoring the pattern to the underlying noise distribution. Finding the optimal pattern corresponds to a high-dimensional optimization problem over the 32-dimensional input vector of the probe search strategy, describing the 16 touch points in the XY-plane. We use \ac{dpse} to automatically optimize the touch points without assuming knowledge about the hole distribution. \ac{dpse} is trained on $M_{train}=1000$ simulated training tasks (hole distributions) with $N_{train}=128$ probe searches each and finetuned on $N_{test}=128$ probe searches over $M_{test}=1$ concrete hole distribution. We compare against the following baselines:
\begin{enumerate}
	\item A fixed 4x4 grid covering the complete search region
	\item A heuristic tailored specifically to probe search, which fits a 16-mode \ac{gmm} to the last points of those trajectories in the finetuning dataset where the search was successful
	\item $\mu+\lambda$ \ac{nsga} with $\mu = \lambda = 100$ (sim) or $\mu = \lambda = 30$ (real)
\end{enumerate}
The experiment is repeated for 10 different multimodal Gaussians $\{H_t\}$ in a simulated environment, and for 6 additional distributions on a real Fanuc industrial robot. Both \ac{dpse} and \ac{nsga} minimize $\mathcal{L}_{fail}(y)$, the probability of search failure. We also assess the effects of adding additional regularization to \ac{dpse}, penalizing either the L1 distance from the initial parameterization ($\mathcal{L}_{init}$, a heuristic-free regularizer) or the smallest distance between any two optimized touch points ($\mathcal{L}_{cdist}$, a heuristic regularizer specific to probe search).

\label{sec:spike_experiment}
\begin{figure*}[ht!]
	\centering
	\subfloat[a]{%
		\includegraphics[width=.15\textwidth]{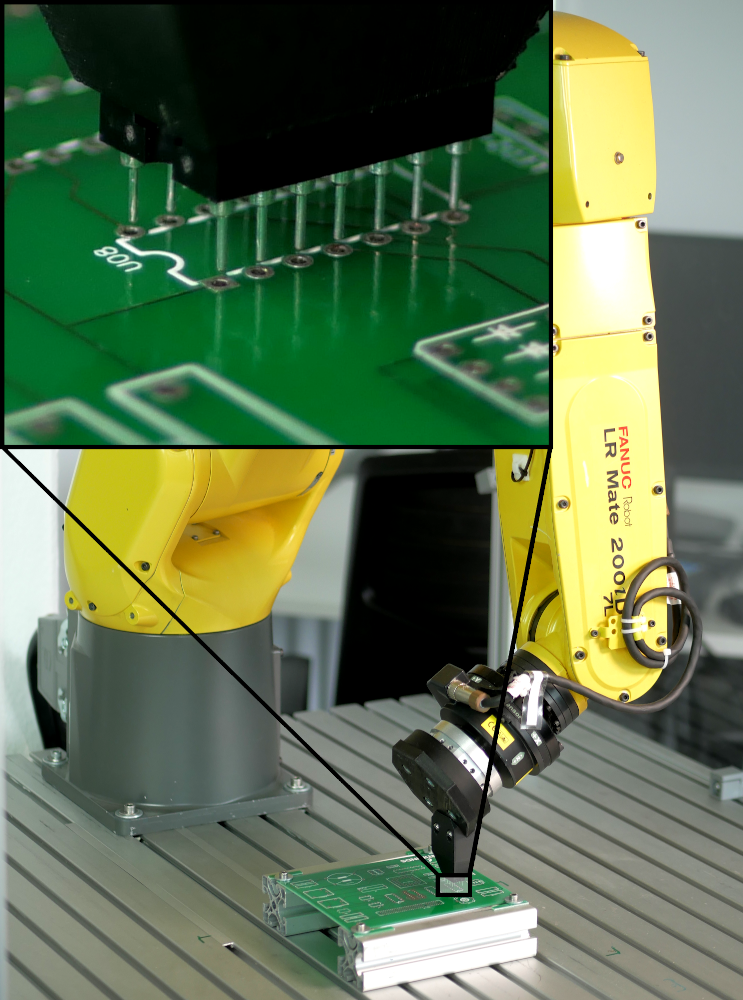}
	}%
	\subfloat[b\label{fig:spike_experiment_results_grid}]{%
		\includegraphics[width=.275\textwidth]{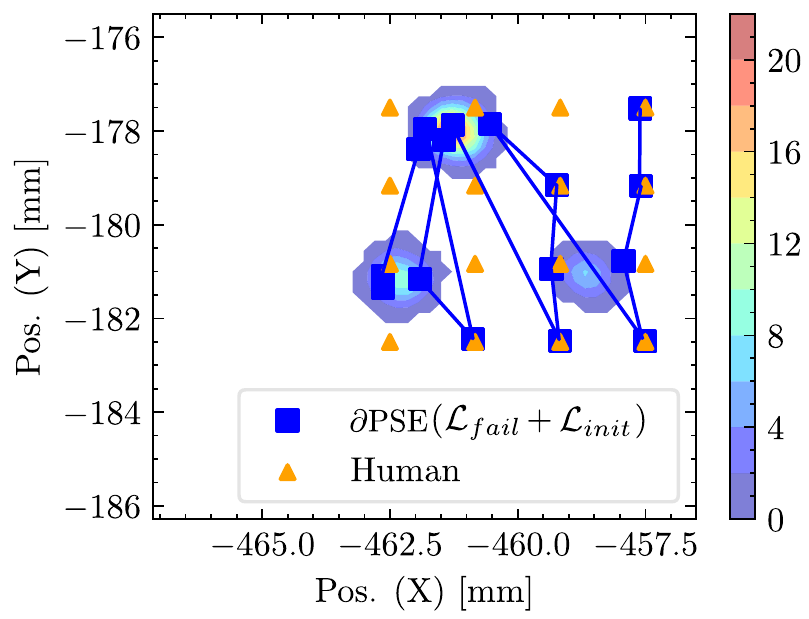}
	}%
	\subfloat[c\label{fig:spike_experiment_results_cdist}]{%
		\includegraphics[width=.275\textwidth]{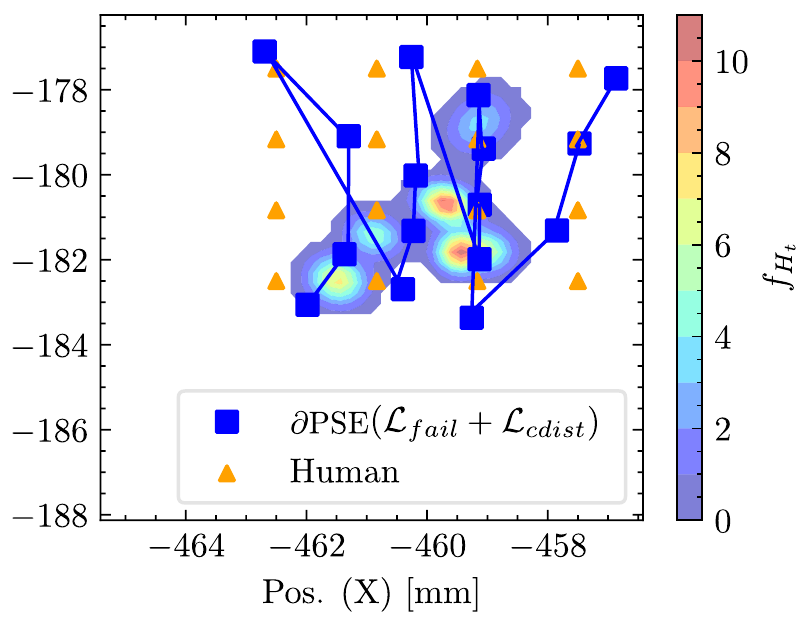}
	}%
	\subfloat[d\label{fig:spike_experiment_results_bars}]{%
		\includegraphics[width=.3\textwidth]{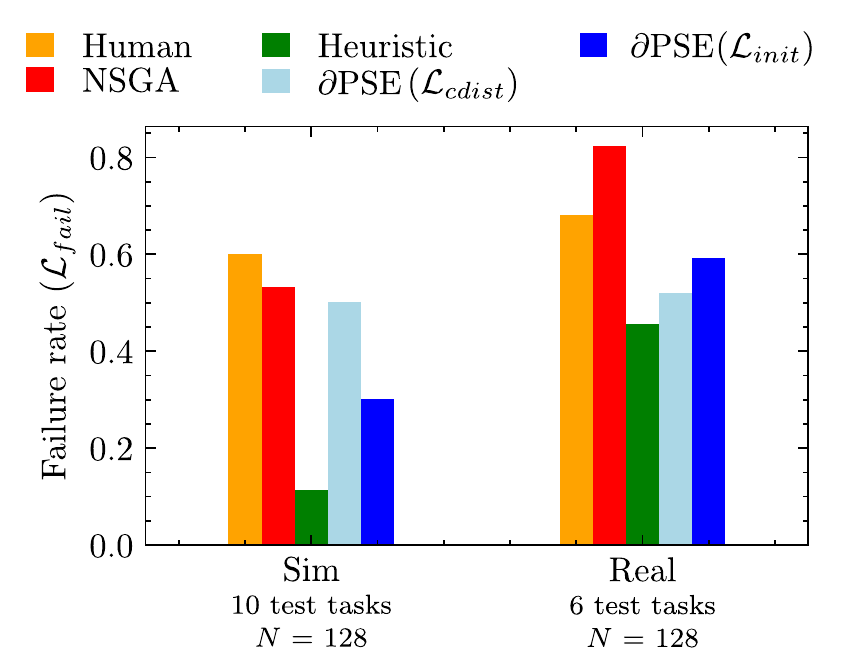}
	}%
	\caption{Results for experiment \ref{sec:spike_experiment} (probe search, stationary): Experiment setup (a) and examples for optimized search patterns generated by \ac{dpse} with the $\mathcal{L}_{init}$ (b) and $\mathcal{L}_{cdist}$ (c) regularizers. Heuristic-free \ac{dpse} outperforms the human and \ac{nsga} baselines and is competitive with an application-specific heuristic (d).}
	\label{fig:spike_experiment_results}
\end{figure*}

The results are summarized in figure \ref{fig:spike_experiment_results}. With both regularizers, the search patterns produced by \ac{dpse} reflect the underlying hole distribution without overfitting to its modes (see fig. \ref{fig:spike_experiment_results_grid} and \ref{fig:spike_experiment_results_cdist}). Quantitatively, both variants of \ac{dpse} outperform the human and \ac{nsga} baselines in both simulated and real environments. While the application-specific heuristic is nearly optimal in the simulated scenario, \ac{dpse} with $\mathcal{L}_{cdist}$ regularization is competitive with it in the real-world environment, while avoiding any assumptions on the underlying hole distributions. The data efficiency of \ac{dpse} is particularly apparent when compared to \ac{nsga}: In the real-world scenario, \ac{nsga} yielded subpar results despite being allowed 300 executions, more than twice \ac{dpse}'s 128. \ac{dpse} achieves this efficiency by transferring knowledge learned from simulated data during pretraining, and by exploiting gradient information during optimization.

\subsection{Probe Search (nonstationary)}
\label{sec:spike_nonstationary_experiment}

In a further series of experiments, we analyze the capacity of \ac{dpse} to optimize search strategies with respect to nonstationary processes. We consider the same task as in experiment \ref{sec:spike_experiment}, but omit the stationarity assumption on $\{H_t\}$. We evaluate \ac{dpse} in a simulated environment on three different nonstationary processes:
\begin{enumerate}
	\item Linear \textit{drift}, where the modes of $\{H_t\}$ are translated by a constant offset at each timestep $t$
	\item \textit{Brownian motion}, where the modes of $\{H_t\}$ are translated by an offset sampled from a bivariate unimodal Gaussian at each timestep $t$
	\item \textit{Shift}, where the modes of $\{H_t\}$ are translated by a uniformly random offset with probability $p_{shift}=0.05$ at each timestep $t$
\end{enumerate}
The results are summarized in table \ref{tb:nonstationary_spikes}. Both unregularized and regularized \ac{dpse} increase the probability of success by over 20\% over short time horizons, confirming the results from experiment \ref{sec:spike_experiment} even as the underlying distributions change over time. For long-horizon processes, the benefits of \ac{dpse} are more pronounced, illustrating the capacity of \ac{dpse} to continuously produce suitable parameterizations over longer timescales.

\begin{figure*}[ht!]
	\centering
	\includegraphics[width=\linewidth]{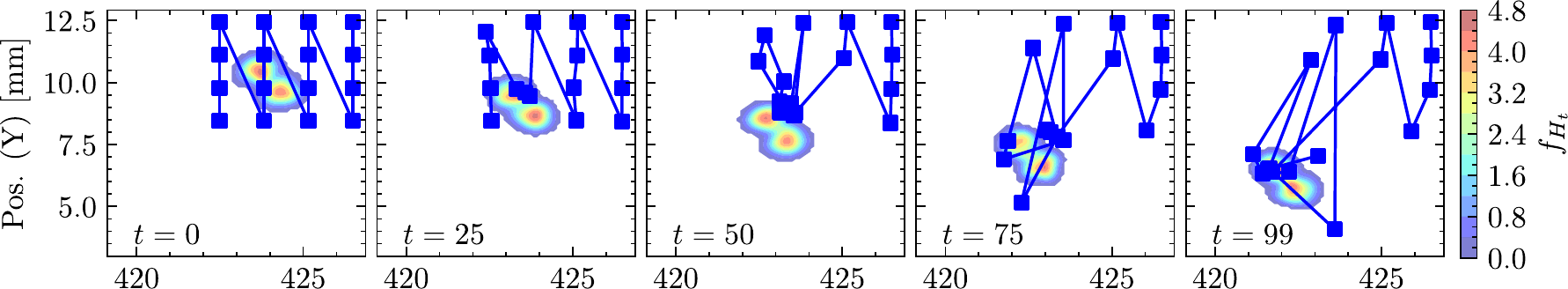}\\
	\vspace{-7.5pt}
	\hrulefill\par
	\includegraphics[width=\linewidth]{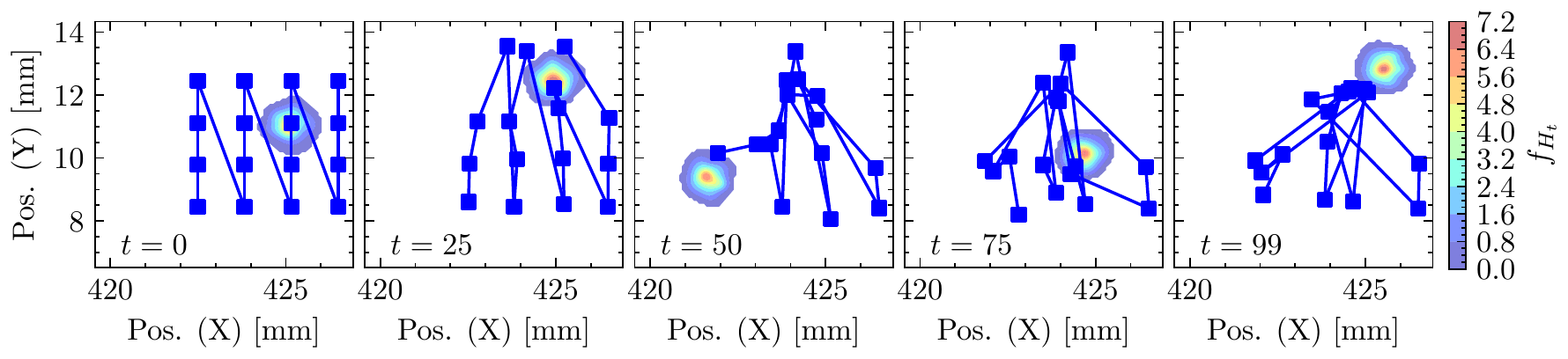}
	\caption{Optimization of a probe search strategy (\ac{dpse}, $\mathcal{L}_{init}$ regularization) in the presence of nonstationary noise processes (top: drift, bottom: brownian motion). For both noise processes, the search pattern (blue) follows the hole distribution, while respecting the constraints imposed by the regularizer.}
	\label{fig:spike_nonstationary}
\end{figure*}

\begin{table}
	\renewcommand{\arraystretch}{1.3}
	\caption{Results of experiment \ref{sec:spike_nonstationary_experiment}: Failure rates for different parameter optimizers and stochastic processes. \ac{dpse} with the $\mathcal{L}_{cdist}$ regularizer reduces failures by up to 53\% over 100 timesteps.}
	\label{tb:nonstationary_spikes}
	\centering
	\input{tables/nonstationary_spike_results.tex}
\end{table}

\subsection{Comparison with Meta-Learning Approaches}
\label{sec:comparison_meta_learning}

\begin{table}
	\renewcommand{\arraystretch}{1.3}
	\caption{Results of experiment \ref{sec:comparison_meta_learning}: Comparison of pretraining/fine-tuning against meta-learning alternatives for training \ac{dpse}'s shadow program.}
	\label{tb:meta_learning_approaches}
	\centering
	\input{tables/comparison_against_meta_learning.tex}
\end{table}

Meta-learning, learning to efficiently learn from few training examples, is a powerful paradigm fundamentally related to \ac{dpse}'s transfer-learning scheme and has become state of the art for solving few-shot learning problems. We find that in the context of gradient-based parameter optimization, simple pretraining and fine-tuning outperforms various state-of-the-art meta-learning approaches. We substantiate our finding by comparing \ac{dpse} as proposed in \ref{sec:tspi} with variants of \ac{dpse}, where the pretraining and fine-tuning of the shadow program is replaced by the meta-learning and adaptation schemes of first- and second-order \ac{maml} \cite{finn_model-agnostic_2017} and Reptile \cite{nichol_first-order_2018}. Comparing against model-agnostic approaches ensures comparability, as the network architectures as well as the training data remain unchanged. Due to the high memory requirements of second-order \ac{maml}, it could only be evaluated for meta-test sets of size $N=5$ (vs. \ac{dpse}'s $N=128$). We evaluate \ac{fomaml} for both $N=5$ and $N=128$ and Reptile for $N=128$.
As shown in table \ref{tb:meta_learning_approaches}, the proposed pretraining and fine-tuning scheme results in a better predictor than all tested meta-learning alternatives. The poor performance of second-order \ac{maml} is likely due to its limitation to only 5 adaptation examples, which do not suffice to learn meaningful characteristics of the underlying distributions. However, the poor performance of \ac{fomaml} on the larger meta-test set indicates that the benefits of meta-learning diminish as the number of meta-test examples increases. This suggests that simpler, much less computationally intensive transfer-learning based approaches can compete with and even outperform meta-learning in the small-data regime (between tens and hundreds of training examples). Our findings also confirm the intuition of Sun et al. \cite{sun_meta-transfer_2019-1} that meta-learning is most effective for shallow network architectures and requires a large amount ($M \sim 100k$) of meta-training tasks. The chained deep residual GRUs at the core of \ac{spi} for precise trajectory prediction and the limited amount of pretraining tasks ($M=1000$) likely limit the utility of meta-learning for gradient-based parameter optimization, at least in conjunction with \ac{dpse}.

\section{Conclusion and Outlook}
\label{sec:conclusion}

In this paper, we propose \acf{dpse}, a novel approach to optimizing robot search strategies. By conditioning learned predictive models of robot skills via transfer learning and embedding them in a differentiable program representation, we obtain a heuristic-free optimizer for arbitrary search strategies. By applying our method to two real-world assembly tasks with stationary process noise and one simulated nonstationary process, we demonstrated that \ac{dpse} outperforms other heuristic-free optimizers and is competitive with task-specific, hand-crafted heuristics. \ac{dpse} permits the effective automation of the labor-intensive optimization phase of robot workcells, while its extreme data-efficiency as well as its black-box nature renders it equally suitable for service robotics tasks. Our future work is focused on extending \ac{dpse} to applications beyond search strategies, such as force-controlled insertion or grasping. In addition, we are exploring the inference of task-specific goal functions from human demonstrations and symbolic knowledge.\pagebreak

\bibliographystyle{IEEEtran}  
\bibliography{bibliography}  

\end{document}

%% file: tables/nonstationary_spike_results.tex
\begin{tabular}{lrrr}
    \hline
    & Drift & Brownian & Shift \\
    \hline
    None &  0.679   & 0.679  & 0.600 \\
    \ac{dpse} ($\mathcal{L}_{fail}$) & 0.458 & 0.605 & 0.472\\
    \ac{dpse} ($\mathcal{L}_{fail} + \mathcal{L}_{init}$) & 0.556 & 0.616 & 0.406 \\
    \ac{dpse} ($\mathcal{L}_{fail} + \mathcal{L}_{cdist}$) & \textbf{0.318}  & \textbf{0.494} &  \textbf{0.339} \\
    \hline
\end{tabular}

%% file: tables/comparison_against_meta_learning.tex
\begin{tabular}{lrr}
    \hline
    Learning algorithm & Traj. loss & Success acc. \\
    \hline
    Pretrain & \textbf{0.80} & \textbf{0.72}  \\
    Pretrain + Dropout & 0.83 & 0.71  \\
    MAML (5) + Dropout & 0.94 & 0.64  \\
    MAML (5) & 0.95 & 0.64  \\
    FOMAML (5) & 0.96 & 0.62  \\
    FOMAML (128) & 0.99 & 0.58  \\
    Reptile (128) & 1.35 & 0.56  \\
    \hline
\end{tabular}